\documentclass{article}
\pdfoutput=1
\usepackage[a4paper, total={6in, 9in}]{geometry}
\usepackage{hyperref}
\usepackage[T1]{fontenc}
\usepackage{lmodern}

\usepackage{url}
\usepackage{booktabs}
\usepackage{framed,multirow}
\usepackage{microtype}
\usepackage{amsmath}
\usepackage{amssymb}
\usepackage{amsfonts}
\usepackage{amsthm}
\usepackage{amsopn}
\usepackage{bm} 
\usepackage{graphicx}
\usepackage{amsfonts}
\usepackage{longtable}
\usepackage{epsf, epsfig}
\usepackage{epstopdf}
\usepackage{subfigure}
\usepackage{caption}
\usepackage{algorithm}
\usepackage[table]{xcolor}
\usepackage{multirow}
\usepackage{diagbox}
\usepackage{footnote}
\usepackage{bbding}
\usepackage{xargs} 
\usepackage{xspace}
\usepackage{diagbox}
\usepackage{xr}
\usepackage{authblk}
\usepackage{paralist}
\usepackage{titling}
\usepackage{nicefrac}
\usepackage{crimson}

\newcommand{\vct}[1]{\boldsymbol{#1}} 
\newcommand{\mat}[1]{\boldsymbol{#1}} 
\newcommand{\field}[1]{\mathbb{#1}}
\newcommand{\R}{\field{R}}

\newcommand{\Map}[1]{\mathcal{#1}}

\newcommand{\ie}{{\emph{i.e.}}}
\newcommand{\eg}{{\emph{e.g.}}}
\newcommand{\draft}[1]{}
\pretitle{\begin{center} \LARGE \bf}
\posttitle{\par\end{center}\vskip 0.5em}
\preauthor{
    \begin{center}
    \normalsize \lineskip 0.5em%
    \begin{tabular}[t]{c}
}
\postauthor{\end{tabular}\par\end{center}}
\predate{\begin{center}\large}
\postdate{\par\end{center}}

\title{Can Deep Learning Outperform Modern Commercial CT Image Reconstruction Methods?}
\author{Hongming Shan, Atul Padole, Fatemeh Homayounieh, Uwe Kruger, Ruhani Doda Khera,\\ Chayanin Nitiwarangkul, Mannudeep K. Kalra, Ge Wang}
\affil{
    Contact: \texttt{\normalsize shanh@rpi.edu, MKALRA@mgh.harvard.edu, wangg6@rpi.edu}
}
\affil{Rensselaer Polytechnic Institute\\Massachusetts General Hospital}

\date{\today}

\begin{document}

\maketitle

\begin{abstract}
Commercial iterative reconstruction techniques on modern CT scanners target radiation dose reduction but there are lingering concerns over their impact on image appearance and low contrast detectability. Recently, machine learning, especially deep learning, has been actively investigated for CT. Here we design a novel neural network architecture for low-dose CT (LDCT) and compare it with commercial iterative reconstruction methods used for standard of care CT. While popular neural networks are trained for end-to-end mapping, driven by big data, our novel neural network is intended for end-to-process mapping so that intermediate image targets are obtained with the associated search gradients along which the final image targets are gradually reached. This learned dynamic process allows to include radiologists in the training loop to optimize the LDCT denoising workflow in a task-specific fashion with the denoising depth as a key parameter. Our progressive denoising network was trained with the Mayo LDCT Challenge Dataset, and tested on images of the chest and abdominal regions scanned on the CT scanners made by three leading CT vendors. The best deep learning based reconstructions are systematically compared to the best iterative reconstructions in a double-blinded reader study. It is found that our deep learning approach performs either comparably or favorably in terms of noise suppression and structural fidelity, and runs orders of magnitude faster than the commercial iterative CT reconstruction algorithms. 
\end{abstract}

\section{Introduction}

Computer vision and image analysis/post-processing are great examples of successes with machine learning especially deep learning. While computer vision and image analysis deal with existing natural images and produce features of these images (images to features), tomographic image reconstruction produces images of internal structures from measurement data, which are various features (line integrals, Fourier/harmonic components, and so on) of the underlying images (features to images). Recently, deep learning techniques have attracted a major attention for tomographic image reconstruction, which is a new area of research \cite{wang2016perspective,wang2018image,zhu2018image}.

Computed tomography (CT) is a widely used imaging modality in biomedical, industrial, and other applications \cite{brenner2007computed}, which utilizes x-ray radiation to create cross-sectional images in a non-invasive manner. Despite overwhelming evidence of healthcare benefits, the extensive use of CT has raised concerns on potential risk of cancer or genetic damage with x-ray radiation \cite{de2009projected,smith2009radiation}. Many clinical indications can be imaged with low-dose CT (LDCT) to minimize the radiation-related risk without significantly compromising the screening or diagnostic performance \cite{national2011reduced}. In fact, decreasing the CT radiation dose as low as reasonably achievable (the ALARA principle) is the commonly accepted practice, and LDCT has been a hot research topic in the medical imaging field for almost two decades. The reduction of radiation dose, however, increases data noise and can introduce artifacts in reconstructed images, which may adversely affect its diagnostic performance if these problems go untreated.

To address this challenge, various noise reduction algorithms were proposed for LDCT, which can be categorized into the following categories: 1) sinogram filtration \cite{wang2005sinogram,manduca2009projection,wang2006penalized}, 2) iterative reconstruction \cite{geyer2015state,willemink2013iterative,zheng2017pwls}, and 3) image post-processing \cite{chen2017cnn,chen2017low,wolterink2017generative,yang2017low}. Sinogram filtration methods process either raw data or log-transformed data prior to image reconstruction. In the data domain, the well-known noise characteristics can be directly utilized to help the design of sinogram filters. However, the resultant methods often suffer from edge blurring or resolution loss, and sinogram data are usually inaccessible to most researchers. Iterative reconstruction methods optimize an objective function that combines the statistical properties of data and the prior information on images. Unfortunately, these iterative techniques are time-consuming, require the sinogram data format, involve hyper-parameters that can only be empirically adjusted, and do not offer consistent image quality improvements \cite{geyer2015state}. Different from the above two kinds of algorithms, image post-processing techniques directly process an image that has already been reconstructed from raw data and is publicly available (subject to the patient privacy, which can be addressed by the IRB approval requesting anonymization of images, etc.). 

The main motivation of this study is to demonstrate whether deep neural networks perform better than the modern commercial iterative reconstruction methods for LDCT and establish a foundation for CT reconstruction algorithms to be empowered by big data and deep learning. For this purpose, our method of choice is a LDCT denoising approach implemented with a novel deep neural network. In addition to the general applicability of this post-processing strategy, the implication is clear that if the post-processing network could already do comparably or favorably than the state-of-the-art iterative algorithms, then the inclusion of machine learning elements in the sinogram domain and the reconstruction process will only further increase the merits of the deep learning approach over the iterative image reconstruction. Also, only with this post-processing approach could we compare the machine learning methods fairly among all the three major CT vendors since all of them do not release raw data formats to us. It is underlined that our intent is to show the superiority of machine learning over the iterative reconstruction algorithms as implemented by the leading industrial CT vendors, instead of comparing among CT image quality metrics among these vendors.

With the rapid development of deep learning techniques, convolutional neural networks (CNNs) have recently achieved state-of-the-art results for LDCT image denoising  \cite{chen2017low,wolterink2017generative,kang2018deep,yang2017low,shan20183d}. Currently, the deep-learning-based methods only learn the end-to-end mapping from LDCT images to NDCT counterparts by minimizing a quantitatively defined loss function. In this context, a conventional loss function may not well reflect radiologists' preference. Here we propose a new paradigm for LDCT imaging via progressive denoising that applies deep learning for end-to-process mapping with radiologists in the loop so that the LDCT imaging process can be effectively and efficiently guided by domain experts in a task-specific fashion. As radiation dose increases from low to high, CT image quality is gradually improved. To a significant degree, this process can be step-wise mimicked through deep learning from a low image quality associated with an order of magnitude less radiation dose to a high image quality of a NDCT reconstruction.  A novel aspect of our approach is to decompose this process into a number of identical network modules, each of which only makes a small increment in image quality and how many improvement steps are needed will be visually judged by a group of experienced radiologists so that the diagnostic performance can be optimized. Interestingly, each network module is analogous to a gradient search step, and the process can cover the conventional end-to-end mapping as a special case, and goes beyond it since the NDCT image is not perfect and subject to further refinement, which can be achieved according to the gradient direction learnt from training data. With the number of denoising steps, which is also referred to as the mapping depth, as the key parameter, the interface between deep learning and radiologists becomes cost-effective and user-friendly, paving the path for machine learning to impact radiology practice proactively. Varying the mapping depth adaptively allows features of interest to be brought into a focus in a fashion not only clinical-task-specific but also image-region-specific, under radiologists' control.

\begin{figure*}[htbp]
	\centering
	\includegraphics[width=1.0\linewidth]{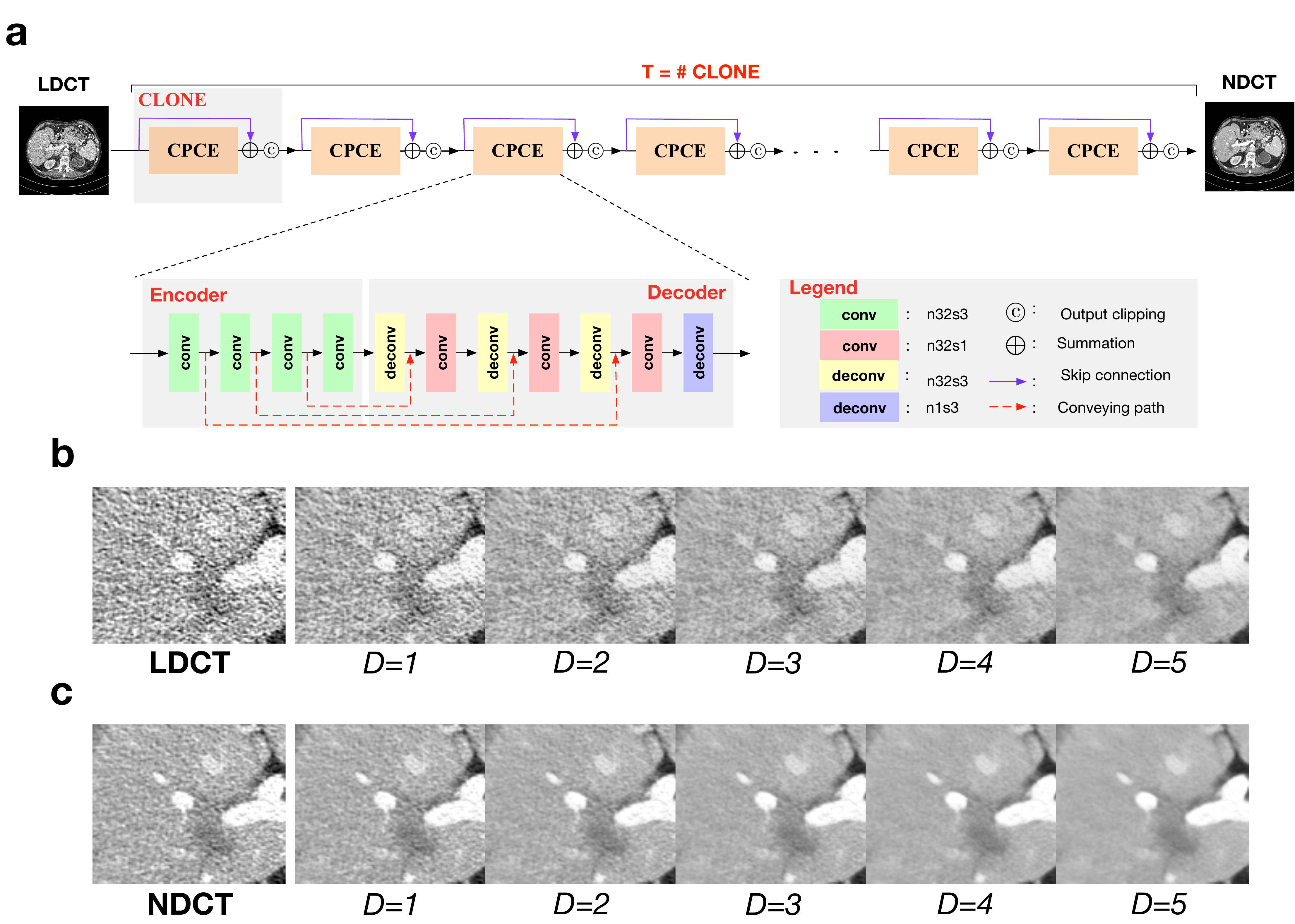}
	\caption{Our proposed Modularized Adaptive Processing (MAP) Neural Network (MAP-NN) for LDCT denoising and the progressively denoised images on the LDCT and NDCT images with $D$ being the mapping depth. (\textbf{a}) Each processing module is a Conveying-Link-Oriented Network Encoder-decoder (CLONE) containing a skip connection to link the input to the output, an Conveying-Path-based Convolutional Encoder-decoder (CPCE) network to learn the residual between its input and output, an summation operation to add the input and residual together, and an output clipping to avoid an exploding gradient. $n32s3$ indicates $32$ filters of size $3\times3$, each (de)convolutional layer is followed by a ReLU; (\textbf{b}) Applying the trained MAP-NN to LDCT images from the Mayo dataset confirms that the MAP-NN can learn the denoising direction from LDCT images to NDCT counterparts, and produce intermediate denoised results that improve the image quality progressively; (\textbf{c}) Applying the trained MAP-NN to NDCT images from the Mayo dataset shows that the MAP-NN model can be used for images acquired at different dose levels to improve the image quality to various degrees. The optimal mapping depth $D$ can be visually judged by a group of radiologists; in other words, with radiologists-in-the-loop the denoised image quality can be optimized in a task-specific fashion.}\label{map-nn}
\end{figure*}

Fig. \ref{map-nn} presents the structure of the proposed Modularized Adaptive Processing Neural Network (MAP-NN) model consisting of multiple Conveying-Link-Oriented Network Encoder-decoders (CLONEs) for LDCT. The MAP-NN network allows progressive denoising operations, which is different from the conventional denoising model that learns from LDCT images to NDCT images directly.  Formally, the MAP-NN can be formulated as follows:
\begin{equation}\label{model:recursive}
	\mat{I}_{\mathrm{den}} = g^T(\mat{I}_{\mathrm{LD}}) = \underbrace{(g\circ g \circ \cdots \circ )g}_{ T=\#g}(\mat{I}_{\mathrm{LD}})   \approx \mat{I}_{\mathrm{ND}}
\end{equation}
where $\mat{I}_{\mathrm{den}}$, $\mat{I}_{\mathrm{LD}}$, and  $\mat{I}_{\mathrm{ND}}$ denote a denoised image, an LDCT FBP image, and an NDCT FBP image, respectively. The operator $\circ$ denotes a functional composition operation, $g$ denotes an CLONE module, $g^t$ denotes the $t$-fold product of the CLONE $g$, and the number of CLONEs for training is denoted by $T$. The parameters of all the CLONEs are shared.

Actually, the module $g$ can be any existing denoising network, such as a fully-connected convolutional network \cite{chen2017cnn,yang2017low}, a convolutional encoder-decoder network with skip connections \cite{chen2017low}, or in this study the Conveying-Link-Oriented Network Encoder-decoder which is an extension to our earlier direct LDCT denoising network \cite{shan20183d} by coupling with skip connection and output clipping and modularizing into a progressive denoising model. The reason for us to select our CLONE is that the conveying links involved in the CLONE make the model more compact and more flexible. The network architecture of the MAP-NN network is further described in the Appendix. We trained the MAP-NN model for $T = 5$ with the Mayo LDCT Challenge dataset under the loss function including the adversarial loss, the mean-squared error, and the edge incoherence. In the testing phase, we used the mapping depth $D = 1, 2, 3$ for all the cases, denoted as DL1, DL2, and DL3 respectively, for radiologists' evaluation. The progressively denoised images obtained from the trained MAP-NN ($T= 5$) are shown in Fig. \ref{map-nn} (b) and (c). 

\section{Material and Method}

\subsection{Patient data} The Human Research Committee of the MGH Institutional Review Board approved this Health Insurance Portability and Accountability Act compliant prospective clinical study. Also, the Human Research Committee of the RPI Institutional Review Board approved use of these patient data. All included patients have written informed consent prior to their participation in the study. In total, $60$ patients datasets were obtained from MGH, half of them undergoing routine abdomen CT exams and the rest routine chest CT exams, on scanners from vendors A, B, and C respectively and proportionally. All CT exams were acquired on one of the three commercial CT scanners from GE Healthcare (Discovery CT750 HD, Waukesha, WI), Philips Healthcare (Brilliance iCT 256, Andover, MA), and Siemens Healthineers (SOMATOM Definition Flash, Germany), in randomized order to protect identity of the scanners. The LDCT image series were acquired immediately (within 5-10 s) after acquisition of their normal dose, clinically indicated CT (NDCT) series. The inclusion criteria were adult patients (age > 18 years), who were hemodynamically stable, able to communicate in English, follow instructions, and hold their breath for at least 10 seconds to avoid motion artifacts. Patients undergoing urgent CT, or with known contrast allergy, women pregnant or planning to become pregnant were excluded from the study. Hemodynamically unstable patients were also excluded. Cross-sectional measurements (anteroposterior and lateral diameters) at mid-slice location were recorded for all patients. Both the NDCT and LDCT were acquired at identical 100-120 kV, 0.9-1.1 beam pitch, wide detector configuration and 0.5 second gantry rotation time. For the LDCT exams, the tube current was reduced to deliver less than 1 mSv radiation dose to the patients for the LDCT image series (DLP for abdomen LDCT $\leq$ 65 mGy$\cdot$cm; DLP for chest CT $\leq$ 70 mGy$\cdot$cm). The section thickness for the abdomen and chest CT were the same as used in our standard of care clinical practice (abdomen CT = 5mm; chest CT = 2.5mm). Radiation doses for chest and abdominal CT are summarized in Table \ref{tab_ctdi}. The inter-vendor differences in radiation doses for NDCT were due to the differences in patient sizes scanned on different CT scanners. For NDCT exams, all scanners automatically adapted the radiation dose by adjusting the tube current with automatic exposure control technique as per our standard of care clinical practice. Thus, in our study LDCT referred to CT radiation dose less than 1 mSv for both the chest and abdomen CT examinations.  

\subsection{Iterative reconstruction techniques}

The sinogram data of LDCT were reconstructed with the corresponding commercial iterative reconstruction (IR) techniques and conventional filtered back-projection (FBP) method. Among different IR settings for each vendor, we asked radiologists to choose three clinical used IR methods before this reader study. The three selected IR methods were randomly renamed as IR1, RI2, and IR3 for each vendor.  The selected IR methods for GE included adaptive statistical iterative reconstruction (ASIR) at strengths of 50 and 70\%, and model-based iterative reconstruction Veo. The selected IR settings for Philips included Idose-level4, IMR-L1-body Routine, and IMR-L1-body-Soft-Tissue (IMR = iterative model reconstruction). For Siemens, the abdomen and chest were reconstructed using different IR methods. The selected IR settings included IRIDIUM (at strengths of 2, 3, and 4) for abdominal imaging and Safire (at strengths of S2, S3, S4) for chest imaging. Thus, image reconstructions were performed using 9 different image reconstructions techniques, and 3 IR methods per patient, from the sinogram data of LDCT of vendors A, B, and C. The NDCT images for all patients were reconstructed using FBP. All image series (n = 5 series per patient (LDCT, NDCT, and three IR reconstructions) $\times$ 60 patients = 300 image series) were used in this study. The LDCT FBP served as the input to our DL method. To avoid the vendor information being identified by file type, all slices are saved in PNG. For the chest scans, we chose 2 slices from the upper 20\% of the chest and the middle of the chest, respectively. For the abdominal scans, we chose 2 slices from the mid liver and the pelvis respectively. This is because these are the areas which are susceptible to noise and artifacts. Therefore, in total 120 cases (2 slices per patient $\times$ 60 patients) were used for image quality evaluation.

\subsection{Subjective image quality evaluation}

\begin{figure*}[htbp]
    \centering
    \includegraphics[width=0.6\linewidth]{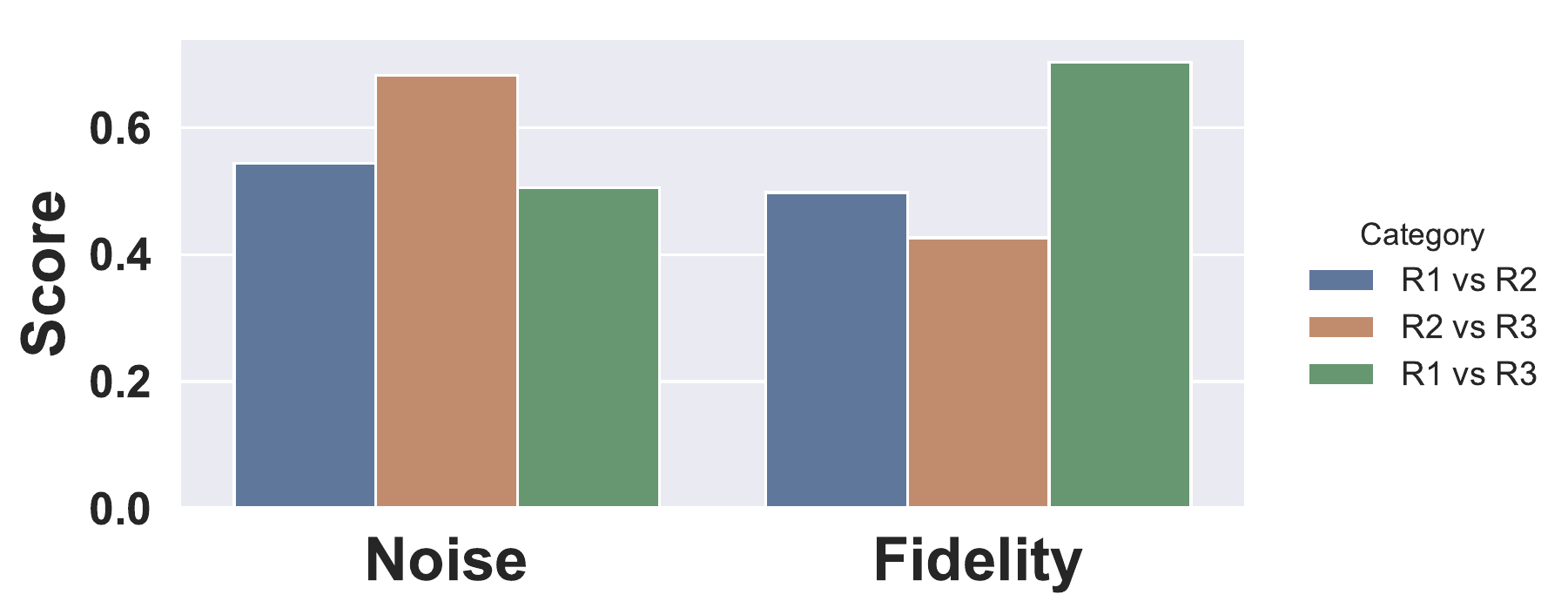}
    \caption{Cohen's kappa statistics for the noise suppression and structural fidelity among three readers on LDCT quality assessment indicating the inter-reader agreement in the range of $[0.42,0.70].$}\label{kappa}
\end{figure*}

Three radiologists (M.K.K with 18 years' experience, C.N. with 5 years' experience, R.D.K. with 4 years' experience) independently evaluated all image cases. Subjective image quality evaluation was performed for each NDCT and LDCT cases independently. The LDCT and NDCT labels were provided to radiologists. Then, three DL and three IR images for all 120 cases of 60 patients were randomized, and independently reviewed in a blinded fashion. The radiologists were asked to assess each image separately for image noise and structural fidelity using a 4-point scale [1= Unacceptable for diagnostic interpretation; 2= Suboptimal, acceptable for limited diagnostic information only; 3 = Average, acceptable for diagnostic interpretation; and 4 = Better than usual, acceptable for diagnostic interpretation]. The radiologists were also asked to comment on any lesions when present. Cohen's kappa statistics for noise and fidelity among three readers on LDCT images shows inter-reader agreement in the range of [$0.42$, $0.70$] as shown in Fig. \ref{kappa}.

\subsection{Data for training MAP-NN}

The training dataset we used is an authorized clinical low-dose CT dataset, which was used for \emph{the 2016 NIH-AAPM-Mayo Clinic Low-Dose CT Grand Challenge}. This dataset included normal-dose abdominal CT images that were taken from 10 anonymous patients and the corresponding simulated quarter-dose CT images.  Poisson noise was inserted into the projection data for each case to reach a noise level corresponding to 25\% of the normal-dose.  For training, 128K image patches of size $64\times64$ were randomly selected from 5 patients that were randomly selected from this dataset. To test the performance of the trained networks, 64K image patches were randomly selected from the remaining 5 patients. The composite loss function and training details can be found in Appendix.

\section{Results}

\subsection{Overall comparison}

We target the overall comparison between deep learning (DL) and iterative reconstruction (IR) in their respective best forms. Each image was evaluated in terms of two aspects: noise suppression which is to compensate for low-dose induced data noise, and structural fidelity which is directly related to the diagnostic performance. Clinically, the structural fidelity is more general and more important than noise level, since the structural fidelity cannot be excellent if the noise level is too high, and DL methods can suppress the noise greatly at a cost of a major structural loss. Thus, given the noise suppression and structural fidelity scores, we first compared the structural fidelity scores, and then checked their noise suppression scores if the structural fidelity scores were indistinguishable. By the nature of this study, we focus on the comparison between the best DL reconstruction and the best IR reconstruction to compare these two competing methodologies. Therefore, the following three hypotheses can be tested:
\begin{itemize}
	\item \textbf{DL > IR}: The best DL reconstruction outperforms the best IR reconstruction in terms of the structural fidelity; or the best DL reconstruction outperforms the best IR reconstruction in terms of noise suppression score when their structural fidelity scores are indistinguishable;
	\item \textbf{DL = IR}: There are no significant differences between the best DL reconstruction and the best IR reconstruction in terms of structural fidelity and noise suppression;
	\item \textbf{DL < IR}: This is the opposite case of the first hypothesis.
\end{itemize}
 
\begin{figure*}[htbp]
	\centering
  \includegraphics[width=1.0\linewidth]{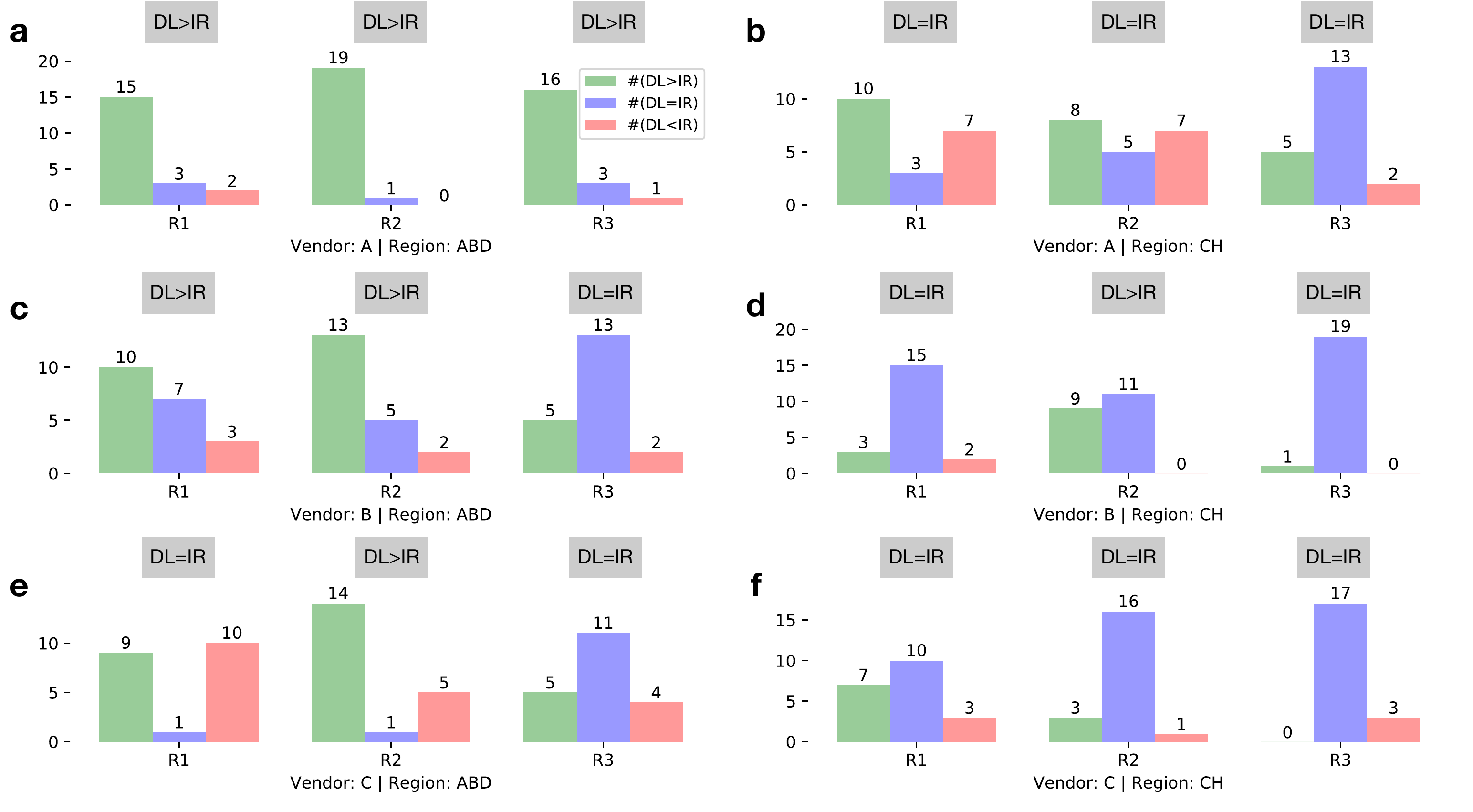}
	\caption{Comparison between the best deep learning (DL) reconstruction and the best iterative reconstruction (IR) for abdomen (ABD) and chest (CH) regions across three major vendors (A, B, and C) and three readers (R1, R2, and R3). (\textbf{a-f}): The histogram showing the number of cases per class (\#(DL>IR), \#(DL=IR), and \#(DL<IR)) in 20 cases on abdomen from vendor A, chest from vendor A, abdomen from vendor B, chest from vendor B,  abdomen from vendor C, and chest from vendor C. The text in the gray box above each plot gives the significant results evaluated by the binomial test at 5\% significant level for each reader. }\label{cmp_all_vendors}
\end{figure*}

Accordingly, we conducted the following two statistical hypothesis tests 
\begin{itemize}
    \item[1)] $H_{0}^{(1)} \ : \ $ \textbf{DL = IR}\qquad $H_1^{(1)} \ : \ $ \textbf{DL > IR}; and
    \item[2)] $H_{0}^{(2)} \ : \ $ \textbf{DL = IR}\qquad $H_1^{(2)} \ : \ $ \textbf{DL < IR}
\end{itemize}
 for a significance of 5\% based on the counts \#(DL > IR) and \#(DL < IR) among $20$ cases, where $\#$ refers to the number of instances where DL outperforms IR or otherwise.  The resultant p-values are reported in Table \ref{pvalue}.  Fig. \ref{cmp_all_vendors} presents the overall comparison between the best DL reconstruction and the best IR reconstruction. Let us highlight the comparison in the following two aspects.

 \textbf{Across CT vendors:} For vendors A and B, all the three readers preferred the best DL reconstruction over the best IR reconstruction for abdominal imaging while the best DL reconstruction was comparable to the best IR reconstruction for the chest scans. For vendor C, DL was comparable to the IR in the abdomen and chest regions. Overall, the conclusion is that the DL method performs better than or comparable to the IR method.

\textbf{Different body regions:} It should be noted that our MAP-NN model was trained on abdomen dataset with the typical abdomen and chest CT windows. After the MAP-NN was trained, we evaluated the model for abdomen and chest regions respectively. In the abdomen cases, all readers rated the DL method better than the IR method on all the vendors except that R1 sometimes rated IR slightly better than DL on vendor C without a statistical significance. Similarly, in the chest cases, all readers rated the DL method better than the IR method on all vendors except that R3  sometimes rated IR slightly better than DL on vendor C without a statistical significance. Overall, the best DL method performs comparably as or better than the best IR method.

\subsection{Performance in noise suppression and structural fidelity}

\begin{figure*}[htbp]
\centering
\includegraphics[height=1.0\linewidth]{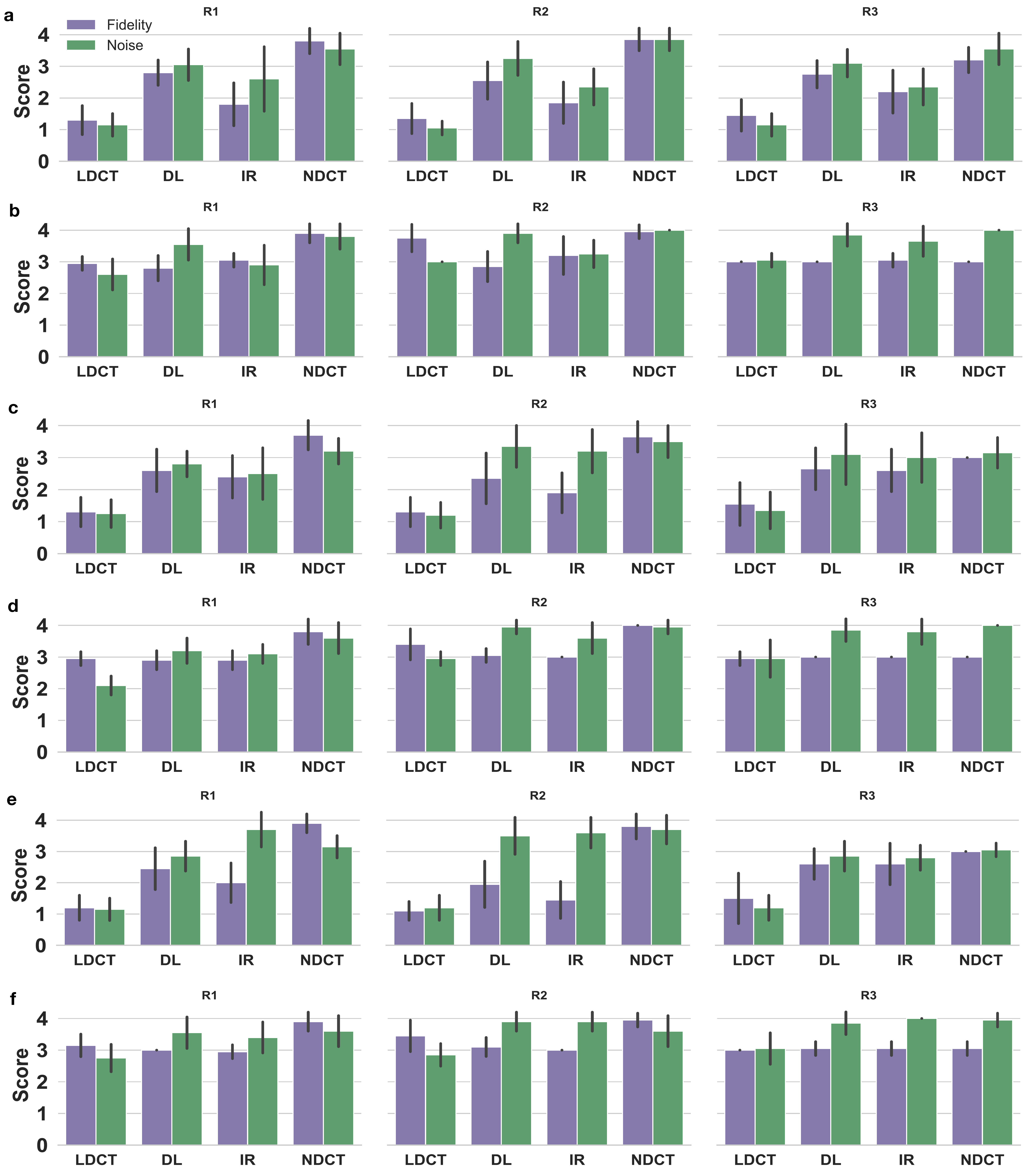}
\caption{Individual metric comparison between the best DL and best IR reconstructions keyed to body regions (Abdomen and Chest) and CT vendors (A, B, and C). (\textbf{a-f}): Average score and standard deviation of the noise suppression and structural fidelity scores on abdomen scans from vendor A, chest scans from vendor A, abdomen from vendor B, chest from vendor B, abdomen from vendor C, and chest from vendor C. The performance metrics are fidelity scores and noise suppression for LDCT, best DL, best IR, and NDCT. In terms of average fidelity scores, DL outperformed IR in 12 of 18 classes (in 3 of the remaining classes DL and IR performed same), while by average noise suppression, DL was superior to IR in 14 of 18 classes.}\label{comparison_FN}
\end{figure*}

Also, we studied the best DL and best IR reconstructions for every selected vendor and each body region in terms of noise suppression and structural fidelity. Fig. \ref{comparison_FN} presents the performance of the best DL and best IR reconstructions in terms of noise suppression and structural fidelity scores. For that purpose, we show the mean scores and standard deviations for noise suppression and structural fidelity respectively. In Fig. \ref{comparison_FN} (a), the DL method achieved a significantly better performance than IR in terms of both scores. In Fig. \ref{comparison_FN} (b), the three readers basically perceived the image quality comparable as obtained using the DL and IR methods. In Fig. \ref{comparison_FN} (c), (d), and (e), DL achieved a better or comparable performance relative to IR. Finally, in Fig. \ref{comparison_FN} (f), all the readers considered that the DL and IR methods were statistically comparable.  As shown in Fig. \ref{cmp_all_vendors}, in some cases, DL and IR gave the same fidelity scores but DL did better in noise suppression. In terms of average fidelity scores, DL outperformed IR in 12 of 18 classes (in 3 of the remaining classes DL and IR performed same), while by average noise suppression, DL was superior to IR in 14 of 18 classes. Therefore, it is clearly concluded that the DL approach performs either comparably or favorably in terms of noise suppression and structural fidelity scores, as compared to the IR approach as implemented by the three leading CT vendors.

\subsection{Sample images from three CT vendors}

\begin{figure*}[htbp]
    \centering
    \includegraphics[width=1.0\linewidth]{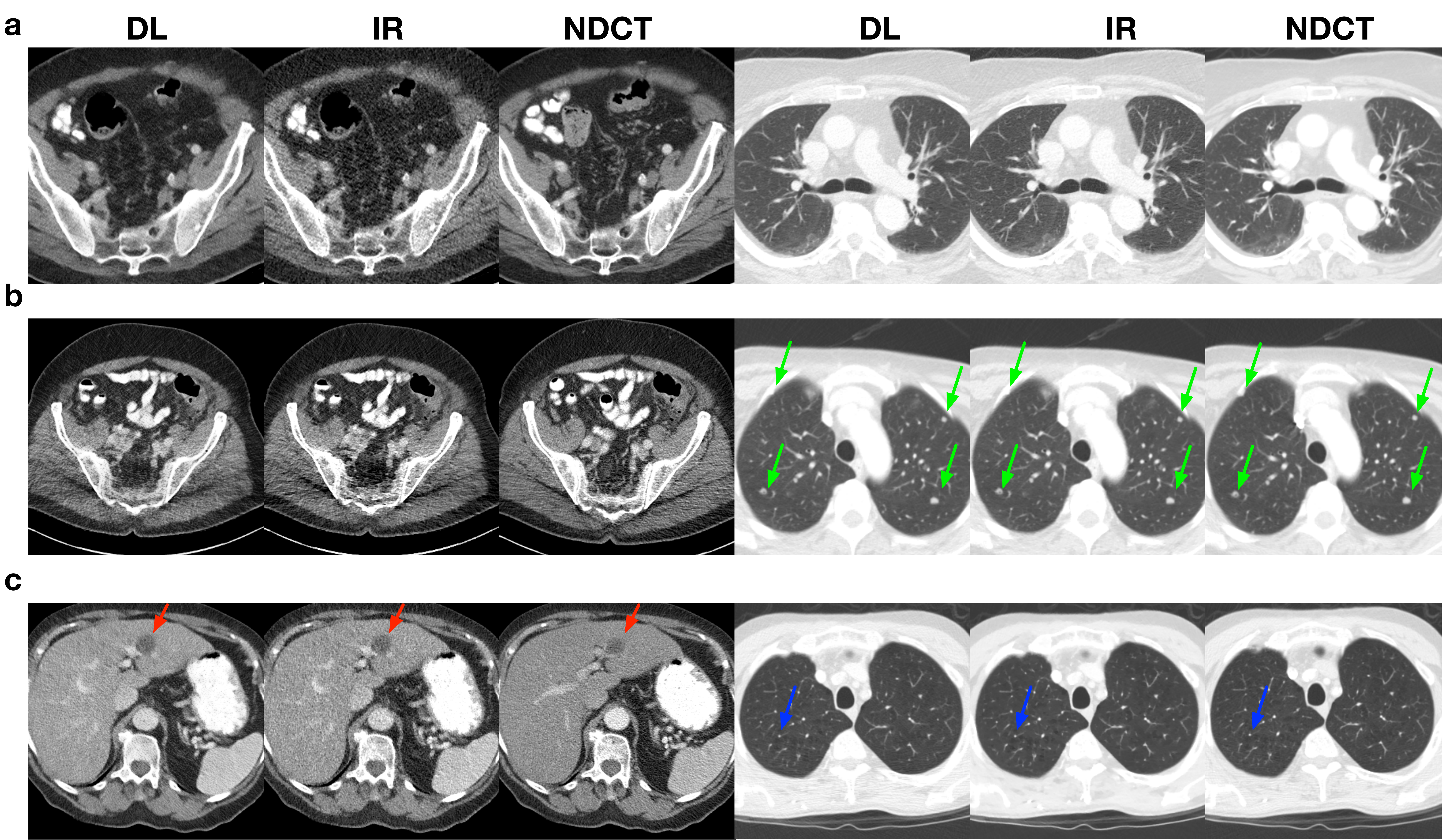}
    \caption{ Best DL and best IR reconstructions as well as NDCT FBP for abdomen and chest regions from three vendors respectively. (\textbf{a-c}): Sample images from vendor A, vendor B, and vendor C respectively. Red, green and blue arrows indicate the liver lesion, lung nodule, and centrilobular emphysema, respectively.}
    \label{sample_img}
\end{figure*}

Fig. \ref{sample_img} presents sample images obtained using the best DL and IR reconstructions respectively from scans on the CT scanners made by the three vendors. The images demonstrate that the DL method for all vendors enabled better noise suppression and structural fidelity than the IR methods. In fact, the IR abdomen CT images (vendors A and B) were deemed unacceptable or limited for noise and fidelity but were acceptable with MAP-NN for all the three vendors.

\subsection{Lesion detectability}

Two of the 30 lesions on NDCT (including a sub-centimeter liver lesion and a tiny apical lung nodule) were not seen on LDCT reconstructed with FBP, IR or DL images. The pseudo-lesion (a focus of enhancement) seen on LDCT FBP and IR method was seen on neither DL nor NDCT images. The remaining lesions (28/30) were seen equally well with IR and DL methods. The liver lesion (red arrows) on abdominal CT images from vendor C in Fig. \ref{sample_img} is equally well seen on NDCT, IR and DL reconstructions. Likewise, four lung nodules (green arrows) on chest CT images from vendor B and centrilobular emphysema (blue arrows) on chest CT images from vendor C are seen on all three image sets (NDCT, IR, and DL), with DL images giving slightly better visibility than IR counterparts.

\section{Discussions}

The proposed MAP-NN consisting of CLONEs allows radiologists-in-the-loop, and performs better than or comparably as the clinically used iterative reconstruction methods implemented by the three leading CT vendors in abdominal and chest regions.  Once our network is trained, DL based denoising is both efficient (about 100 slices per second per mapping depth) and versatile for clinical use, while iterative reconstruction techniques are time-consuming and may introduce significant artifacts.

Compared to the previously published deep-learning-based denoising networks \cite{shan20183d,yang2017low,chen2017low,chen2017cnn,wolterink2017generative} that learn the denoising mapping from images collected at a specific low-dose setting to the NDCT counterparts, our MAP-NN can be viewed as a significant refinement and a major extension, which learns not only intermediate denoised images through multiple CLONE stages but also the associated gradient directions.  Then, the number of CLONE modules, also known as the mapping depth, becomes a key parameter, over which the radiologists have the best judgement on the selection of an optimal mapping depth in a task specific fashion. The MAP-NN with CLONEs permits a cost-effective and user-friendly interface between deep learning and radiologists, enabling the mixed/augmented intelligence beyond what standalone deep learning could achieve. In the Appendix, we give on the differences between the conventional denoising model and the proposed progressive denoising model in more details.

For the first time, our MAP-NN systematically demonstrates that the DL approach can provide a similar or better image quality in terms of structural fidelity and noise suppression as compared to the state-of-the-art commercial IR methods that are based on image reconstruction directly from raw data. Also, the DL approach is much more computationally efficient after offline training. Therefore, the DL approach can already effectively compete with the IR solutions, and potentially replace IR methods. Furthermore, because DL methods can be vendor agnostic, institutions with CT from more than one vendors can anticipate similar image appearance as opposed to what the commercial IR techniques offer. Currently, unique changes in the image appearance are associated with vendor-specific reconstruction programs. This is an obstacle for large-scale radiomics studies, and could be streamlined using DL techniques in the future.

However, there are some limitations of this study. First, as an overall comparative study, the MAP-NN has not been optimized to either a specific vendor or a particular body region. The collection of more cases in the future will help improve the denoising performance and enhance the statistical significance of the denoising gain over the IR results. Second, LDCT and NDCT slices are not in perfect registration, which might affect the evaluation scores to some degree. As a further topic, we can utilize semi-supervised or unsupervised learning techniques for potential image quality refinement. Finally, our DL method was selected to be applicable to CT scans from all these vendors, from all of which we cannot have access to raw data. As a result, more powerful DL methods cannot be implemented without the data format. Despite these limitations, our overall conclusion has been encouraging that DL is either better than or comparable to IR. Clearly, it is now the time for the CT field to open the data format, go machine learning, and develop the next
generation of CT image reconstruction algorithms in the deep learning framework.

In conclusion, our DL method provides better or comparable image quality compared to commercial IR techniques from three CT vendors, and there is great potential for optimization of DL CT reconstruction methods that handle sinogram data directly.

\section{Acknowledge}
The authors would thank Drs. Michael Vannier and Bruno De Man for helpful discussions, and NVIDIA Corporation for the donation of GPU used for this research.

\bibliographystyle{unsrt}
\bibliography{ref}

\newpage
\appendix
\section*{Appendix}

\section{Modularized Adaptive Processing Neural Network (MAP-NN) for Progressive LDCT Denoising}

\subsection{Conventional denoising model}

\begin{figure*}[htbp]
    \centering
    \includegraphics[width=1\linewidth]{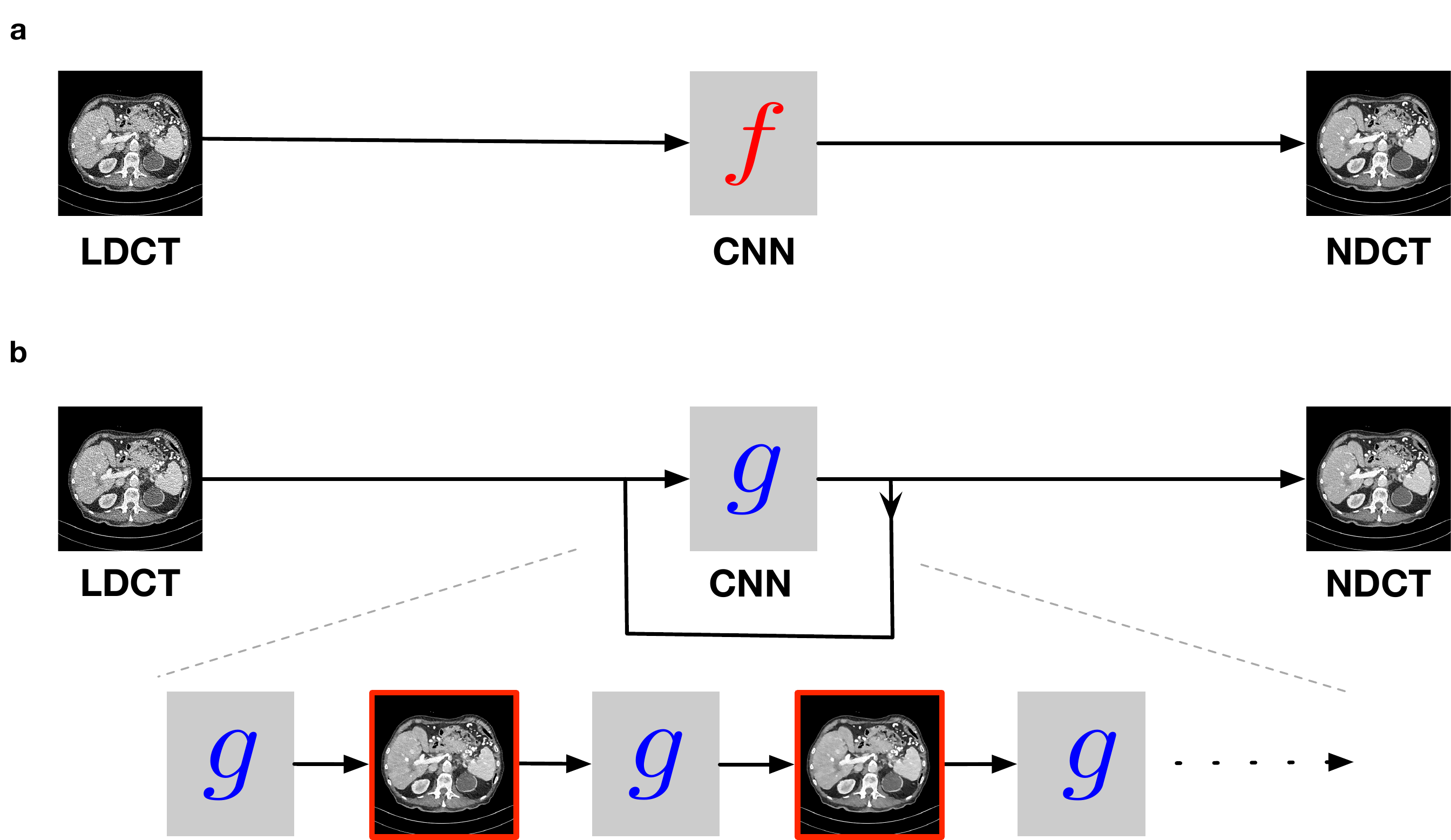}
    \caption{Comparison between the conventional direct denoising model and the proposed progressive denoising model. (\textbf{a}) The conventional direct denoising model aims at learning a direct mapping $f$ from LDCT to NDCT, and (\textbf{b}) our proposed progressive denoising model consists of multiple identical module $g$ to learn the mapping from LDCT to NDCT. The denoised images in the red boxes are the intermediate denoised results along the noise reduction direction. Clearly, the progressive denoising model can reduce to the direct denoising model if there is only one module in the progressive denoising model.}\label{denoising_model}
\end{figure*}

Assume that $\mat{I}_{\mathrm{LD}}\in\R^{w\times h}$ is a low-dose CT (LDCT) image of size $w\times h$,  $\mat{I}_{\mathrm{ND}}\in\R^{w\times h}$ is the corresponding normal-dose CT (NDCT) image, and the relationship between the LDCT and NDCT images can be expressed as follows:
\begin{equation}
	\mat{I}_{\mathrm{LD}} = \Map{N}(\mat{I}_{\mathrm{ND}})
\end{equation} 
where $\mathcal{N}:\R^{w\times h} \rightarrow \R^{w\times h}$ denotes the noise-induced corrupting process due to the quantum nature of the x-ray photon propagation. The conventional denoising model is to provide an approximate inverse function $f\approx \Map{N}^{-1}$ in order to estimate the NDCT image $\mat{I}_{\mathrm{ND}}$ from the LDCT image $\mat{I}_{\mathrm{LD}}$, \ie,
\begin{equation}\label{model:plain}
	\mat{I}_{\mathrm{den}} = f(\mat{I}_{\mathrm{LD}})   \approx \mat{I}_{\mathrm{ND}}
\end{equation}
where $\mat{I}_{\mathrm{den}}$ is the denoised LDCT image. Fig. \ref{denoising_model}  (a) presents the diagram of the conventional denoising model. Different deep-learning-based denoising network structures were proposed to achieve the denoising performance, including fully-connected convolutional networks \cite{chen2017cnn,yang2017low,you2018structurally}, a convolutional encoder-decoder network with skip connections \cite{chen2017low}, a convolutional encoder-decoder with conveying-paths \cite{shan20183d}, and their 3D variants  \cite{wolterink2017generative,shan20183d}.

Despite encouraging denoising results, the conventional denoising model has the following drawbacks. On one hand, as shown in Fig. \ref{denoising_model} (a), the conventional denoising model aims at learning a direct mapping function $f$ from a specific low dose level to the specific normal dose level; \eg, from quarter-dose to normal-dose images in the Mayo Clinic Low-dose CT dataset\footnote{\url{http://www.aapm.org/GrandChallenge/LowDoseCT/}}. Thus, the trained denoising model
may not generalize very well to a new dose level. In practice, a commercial scanner may produce CT scans at different radiation doses according to various protocols. One of the most effective ways to adjust the radiation dose level is to change the tube current. This implies that, in order to deal with different radiation doses, one has to train many denoising models from different radiation doses to the normal dose, which is cumbersome, inflexible, time-consuming, and impractical. On the other hand, the quality of the NDCT image also affects the denoising performance. Different from the image denoising tasks in computer vision, which usually have the noise-free or nearly-noise-free ground truth, a NDCT image, served as the ground truth for training  the denoising networks, still contains substantial noise. Such imperfect NDCT images may compromise the denoising performance to some extent.

\subsection{Progressive denoising model} Our proposed progressive denoising model MAP-NN is a refinement and an extension to the conventional direct denoising model, being capable of addressing the above-mentioned drawbacks. Fig. \ref{denoising_model} (b) presents the diagram of the progressive denoising model, which can be formulated in a modularized format:
\begin{equation}\label{progressive}
	\mat{I}_{\mathrm{den}} = g^T(\mat{I}_{\mathrm{LD}}) = \underbrace{(g\circ g \circ \cdots \circ )g}_{ T=\#g}(\mat{I}_{\mathrm{LD}})   \approx \mat{I}_{\mathrm{ND}}
\end{equation}
where the operator $\circ$ denotes a function composition, $g$ denotes the progressive module called a Conveying-Link-Oriented Network Encoder-decoder (CLONE), $g^t$ denotes the $t$-fold product of the $g$, and the number of CLONEs for training is denoted by $T$. Compared to the conventional denoising model in \eqref{model:plain}, the progressive denoising model MAP-NN in  \eqref{progressive} has the following desirable properties. First, the MAP-NN reduces to the conventional denoising model if there is only one CLONE in the denoising model; \ie, $T=1$. When $T>1$, the progressive denoising model increases the model complexity/depth without adding new parameters \cite{kim2016deeply,tai2017image}, and increases the size of the receptive field as well. This implies that the MAP-NN is a refinement and an extension of the existing denoising model. Second, as shown in the red bounding box of Fig. \ref{denoising_model} (b), the progressive denoising  model  can  produce  a  sequence  of  intermediate  denoised  images  from  each  CLONE,  \ie,  $\{g^i(\mat{I}_{\mathrm{LD}})\}_{i=1}^T$.  Such a sequence of intermediate denoised images can be viewed as a dynamic denoising process, and is directly proportional to a computational elevation of radiation dose. Third, the MAP-NN does not hypothetically require noise-free normal-dose CT images as the labels/targets, since each CLONE seeks removing part of noise from its input. Last and most importantly, for a new dose level one can apply the trained denoising CLONE multiple times, then obtain a sequence of denoised images, and let domain experts select the best denoising result; in other words, with radiologists-in-the-loop the denoised image quality can be optimized in a task-specific fashion even if exact imaging and protocol knowledge is unknown.

\section{MAP-NN architecture}

\subsection{Network structure of MAP-NN}

MAP-NN consists of multiple identical CLONE module in the network. The parameters of CLONEs are shared. The CLONE can be any existing denoising network. The CLONE this study used is an extension to our earlier direct LDCT denoising network Conveying-Path-based Convolutional Encoder-decoder (CPCE) \cite{shan20183d} that was modified for progressive denoising model. The CPCE network has $4$ convolutional layers, each of which has $32 $ filters of size $3\times3$, followed by $4$ deconvolutional layers also with $32$ filters of size $3\times3$, except for the final layer that has only $1$ filter. The filter stride is set to $1$ for all convolutional and deconvolutional layers. The conveying path, originally introduced in the U-net \cite{ronneberger2015u} for biomedical image segmentation, copies the early feature-maps and reuses them as the input to a later layer of the same feature-map size in a network. This mechanics  preserves details of the high-resolution features. Remarkably, the densely-connected convolutional network (DenseNet) \cite{huang2016densely} receives the feature-maps of all preceding layers, achieving the state-of-the-art classification performance on ImageNet. Our denoising CPCE network has three conveying paths, copying the output of an early convolutional layer and reusing it as the input to a later deconvolutional layer of the same feature-map size. To reduce the computational cost, one convolutional layer with $32$ filters of size $1 \times 1$ is used after every conveying path, reducing the number of feature-maps from $64$ to $32$. Each convolutional or deconvolutional layer is followed by a rectified linear unit (ReLU).
 
\subsection{Conveying-Link-Oriented Network Encoder-decoder (CLONE)}

While the deep progressive denoising model is powerful compared to the conventional denoising model, training a deep progressive denoising network with above CPCE module, however, can be very difficult. Among many issues, a major problem is the exploding/vanishing gradients \cite{pascanu2013difficulty}. Another problem is that storing an exact copy of information of many layers is not easy. In LDCT denoising, NDCT images are quite similar to LDCT counterparts, and processing modules need to keep the exact copy of input images for many modules. To address these problems, in this study we improved the CPCE design by incorporating the output clipping and the residual skip connection, forming the CLONE unit for progressive denoising model. Because the output of each CPCE is an image, which can be viewed as the ``bottleneck'' in the progressive process. Therefore, we clip the output into range of $[0,1]$, the same as that of the input range, to prevent gradients from going to extremes and out of control. To avoid vanishing gradients, we use a residual skip connection from the input to the output of each module \cite{tai2017image}. Hence, each module infers the noise distribution instead of the whole image, compressing the output space and avoiding the use of the exact copy of the input image.

\subsection{Discriminator structure}

The discriminator used for optimizing the MAP-NN has $6$ convolutional layers with $64$, $64$, $128$, $128$, $256$, and $256$ filters of size $3\times3$, followed by $2$ fully-connected layers of sizes $1024$ and $1$, respectively. Each convolutional layer is followed by a leaky ReLU, which has a
negative slope of $0.2$ when the unit is saturated and not active. A unit filter stride is used for oddly indexed convolutional layers, and this stride is doubled for evenly numbered layers.

\section{Loss functions}

The composite function for optimizing the network includes three components: adversarial loss, mean-squared error, and edge incoherence. 

\subsection{Adversarial loss}

Recently, the GAN~\cite{goodfellow2014generative} architecture was developed to model the distribution of given data. A GAN has a pair of neural networks $(G, D)$, where $G$ and $D$ are called the Generator network and the Discriminator network, respectively.  It is actually a game  between these two networks, where the Generator learns to produce more and more realistic samples, and the Discriminator learns to become smarter and smarter at distinguishing generated data from ground-truth samples. The two networks are trained alternatively, and the purpose is that the competition drives the generated samples to be hardly indistinguishable from ground-truth. 

In this paper, we used a variant of the original GAN, called the Wasserstein GAN (WGAN) with a gradient penalty \cite{gulrajani2017improved}, which addresses the deficiencies such as low quality of generated images, slow convergence, and mode collapse. For LDCT denoising, the objective function of WGAN can be described as follows~\cite{gulrajani2017improved}:

\begin{align}\label{wgan}
	\min_{\vct{\theta}_G} \max_{\vct{\theta}_D} \Bigg\{  \underbrace{\mathbb{E}_{\mat{I}_{\mathrm{LD}}}\Big[D\big(G(\mat{I}_{\mathrm{LD}})\big)\Big] - \mathbb{E}_{\mat{I}_{\mathrm{ND}}}\Big[D(\mat{I}_{\mathrm{ND}})\Big]}_{\mathrm{\bf Wasserstein~distance}} +  \lambda_{g} \underbrace{\mathbb{E}_{\mat{\bar{I}}}\Big[\big(\| \nabla D(\mat{\bar{I}}) \|_2 - 1\big)^2\Big]}_{\mathrm{\bf gradient~penalty}} \Bigg\},
\end{align} 
where $\mathbb{E}_a[b]$ denotes the expectation of $b$, as a function of $a$, $\vct{\theta}_G$ and $\vct{\theta}_D$ indicate the parameters of networks $G$ and $D$ respectively, $\vct{\bar{I}} = \epsilon\cdot G(\mat{I}_{\mathrm{LD}}) + (1-\epsilon)\cdot \mat{I}_{\mathrm{ND}}$ with $\epsilon$ uniformly sampling from the interval of $[0,1]$.  $\nabla D(\mat{\bar{I}})$ denotes the gradient of $D$ with respect to variable $\mat{\bar{I}}$, the parameter $\lambda_g$ controls the trade-off between the Wasserstein distance and the gradient penalty term, which was suggested to set to $10$ \cite{gulrajani2017improved}. The literature has suggested to determine the optimum of \eqref{wgan} by iteratively optimizing the generator $G$ once and discriminator $D$ four times~\cite{goodfellow2014generative,arjovsky2017wasserstein,gulrajani2017improved}, which we used in this project.

The adversarial loss encourages the generator network to produce samples that are indistinguishable from the NDCT images, which refers to the loss function of the generator in~\eqref{wgan}~\cite{gulrajani2017improved}:
\begin{equation}\label{ad}
	\min_{\vct{\theta}_{G}}\mathcal{L}_a=\mathbb{E}_{\mat{I}_{\mathrm{LD}}}\Big[D\big(G(\mat{I}_{\mathrm{L}})\big)\Big],
\end{equation}
since the last two terms in~\eqref{wgan} are constant with respect to $\vct{\theta}_{G}$. Note that the proposed MAP-NN is the generator $G$ in the WGAN framework. 

\subsection{Mean-squared error}

The mean-squared error (MSE) measures the difference between the output and NDCT images, which would reduce the noise in the input LDCT image. Formally, the MSE is defined as follows:
\begin{equation}\label{mse}
	\min_{\vct{\theta}_{G}} \mathcal{L}_{m}= \mathbb{E}_{(\mat{I}_{\mathrm{LD}},\mat{I}_{\mathrm{ND}})} \Big\| \mat{I}_{\mathrm{ND}} - G(\mat{I}_{\mathrm{LD}})\Big\|^2.
\end{equation} 

\subsection{Edge incoherence}

The Sobel filter is a discrete differentiation operator, computing an approximation of the gradient of the image intensity function. At each point in the image, the result of the Sobel-Feldman operator is either the corresponding gradient vector.  The Sobel-Feldman operator is based on convolving the image with a small, separable, and integer-valued filter in the horizontal and vertical directions, and therefore relatively inexpensive.  As a result, the gradient approximation that it produces is relatively crude, in particular for high-frequency variations in the image. We define the edge incoherence to measure the difference between the Sobel filtrations of real and estimated images as
\begin{equation}\label{sobel}
	\min_{\vct{\theta}_{G}} \mathcal{L}_{e}= \mathbb{E}_{(\mat{I}_{\mathrm{LD}},\mat{I}_{\mathrm{ND}})} \Big\| SF(\mat{I}_{\mathrm{ND}}) - SF(G(\mat{I}_{\mathrm{LD}}))\Big\|^2
\end{equation}
where $SF$ denotes the Sobel filteration. This could enhance the edge information in the denoised image.

\subsection{Final objective function}
The final objective function for minimizing generator is defined as 
\begin{equation}
    \min_{\vct{\theta}_{G}} \mathcal{L} = \mathcal{L}_{a}+\lambda_{m}\mathcal{L}_{m}+\lambda_{e}\mathcal{L}_{e},
\end{equation}
which encourages the generated denoised image to preserve more texture information, reduce the noise, and enhance the edge. Both $\lambda_{m}$ and $\lambda_{e}$ are set as $50$ in the experiments.

\section{Training details}

\begin{figure*}[htbp]
    \centering
    \includegraphics[width=0.45\linewidth]{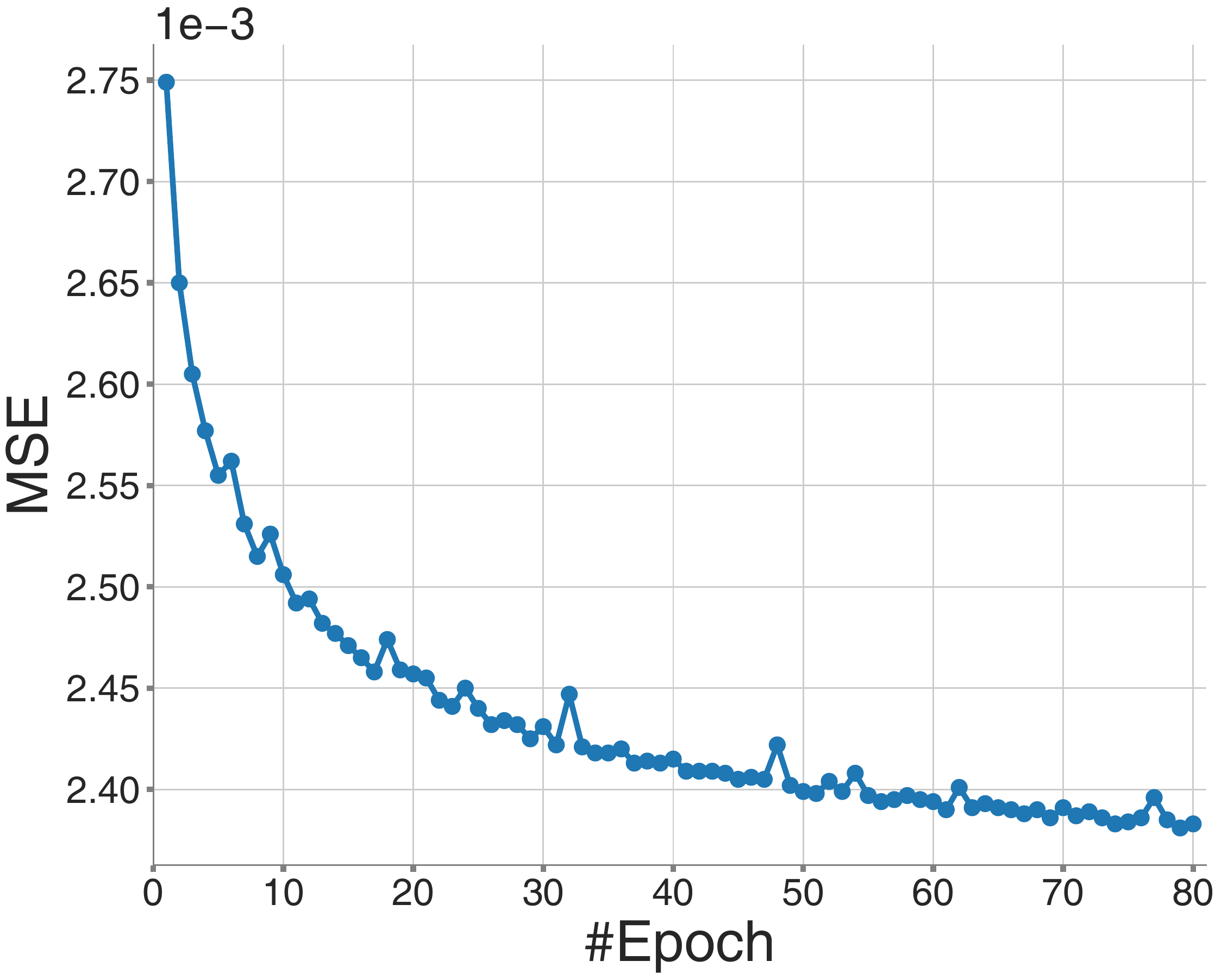}
    \includegraphics[width=0.45\linewidth]{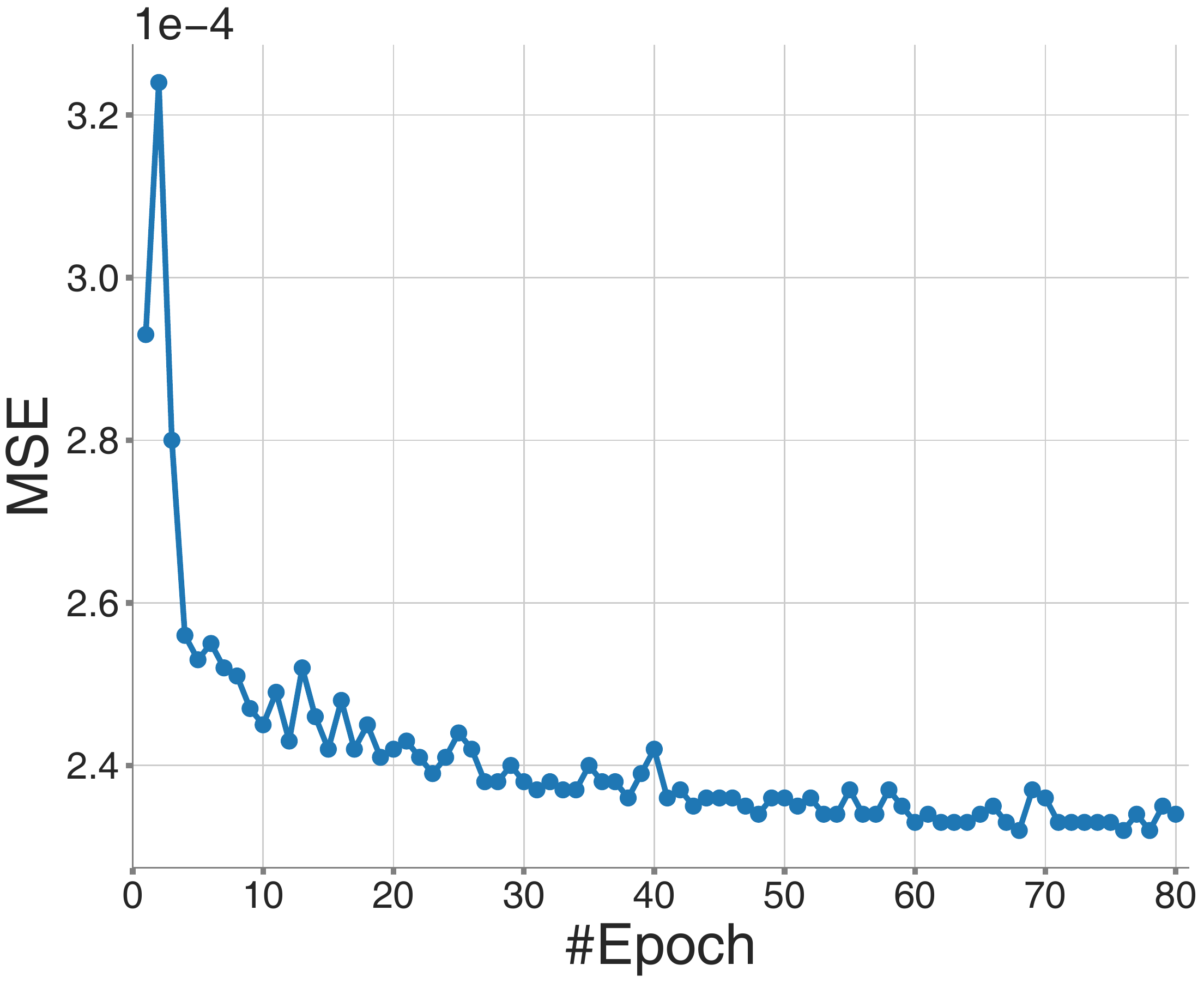}
    \caption{Validation curves for abdomen and chest windows on 64K image patches from Mayo clinical data. Left: Abdomen. Right: Chest. The networks were trained 80 epochs and converged. }\label{validation_curves}
\end{figure*}

The network was optimized using the Adam optimization method \cite{kingma2014adam} with a min-batch of 128 image patches for each iteration.  The learning rate was set to $1.0\times10^{-4}$ with two exponential decay rates $\beta_1=0.9$ and $\beta_2=0.999$ for the moment estimates. The learning rate was adjusted by $1/\sqrt{t}$ decay after every epoch. The network was trained 80 epochs within 24 hours on a NVIDIA 1080Ti GPU. The networks were implemented with deep learning library TensorFlow \cite{abadi2016tensorflow}.  Two testing curves are shown in Fig. \ref{validation_curves} for abdomen and chest CT windows, respectively.

\newpage

\begin{table*}[htbp]
\caption{Mean and standard deviation of CT dose index volume (CTDI\textsubscript{vol}), dose length product (DLP), and estimated effective dose (EED) for normal-dose CT (NDCT) and low-dose (LDCT) for vendor A, B, and C. }\label{tab_ctdi}
\centering
    \begin{tabular}{cccccccc}
    \toprule
    & & \multicolumn{2}{c}{Vendor A} & \multicolumn{2}{c}{Vendor B} & \multicolumn{2}{c}{Vendor C} \\  \cmidrule{3-8}
    & & NDCT & LDCT & NDCT & LDCT & NDCT & LDCT \\  \cmidrule{1-8}
    \multirow{3}{*}{Abdomen CT}  & CTDI\textsubscript{vol}, mGy & $10 \pm 3.8$ & $1.4 \pm 0.3$ & $7 \pm 2.6$ & $1.3 \pm 0.1$ & $8 \pm 2$ & $1.3 \pm 0.3$ \\ 
    &  DLP, mGy$\cdot$cm & $467 \pm 222$ & $64 \pm 2.1$ & $308 \pm 125$ & $59 \pm 1.8$ & $376 \pm 150$ & $61 \pm 3$ \\
    &  EED, mSv &  $7 \pm 3.3$ & $0.9 \pm 0.1$ & $4.6 \pm 1.9$ & $0.9 \pm 0.1$ & $5.6 \pm 2.2$ & $0.9 \pm 0.1$ \\
    \midrule
    \multirow{3}{*}{Chest CT} & CTDI\textsubscript{vol}, mGy & $10 \pm 7$ & $1.8 \pm 0.2$ & $7 \pm 2.5$ & $1.9 \pm 0.2$ & $6 \pm 0.5$ & $1.9 \pm 0.2$ \\ 
    & DLP, mGy$\cdot$cm & $378 \pm 239$ & $66 \pm 3.3$ & $228 \pm 101$ & $62 \pm 1.4$ & $218 \pm 34$ & $67 \pm 2.2$ \\
    & EED, mSv &  $5.3 \pm 3.4$ & $0.9 \pm 0.1$ & $3.2 \pm 1.4$ & $0.9 \pm 0.1$ & $3.1 \pm 0.5$ & $0.9 \pm 0.1$ \\
    \bottomrule
    \end{tabular}
\end{table*}

\begin{table*}[htbp]
\caption{P-values for the hypothesis $H_0^{(1)}$ and $H_0^{(2)}$ across three vendors (A, B, and C), two body regions (Abdomen and Chest), and three readers (R1, R2, and R3). The blue cells indicate that we need to reject $H_0^{(1)}$ at a 5\% significant level; in other words, the best deep learning reconstruction is better than the best iterative reconstruction. In the remaining cells, we do not have sufficient evidence to reject $H_0^{(1)}$ or $H_0^{(2)}$; in other words, the best deep learning reconstruction is comparable to the best iterative reconstruction.}\label{pvalue}
\centering
    \begin{tabular}{cccccccc}
    \toprule
    \multirow{2}{*}{Vendor}& \multirow{2}{*}{Region}& \multicolumn{2}{c}{R1} & \multicolumn{2}{c}{R2} & \multicolumn{2}{c}{R3} \\ \cmidrule{3-8}
    & & $H_0^{(1)}$ & $H_0^{(2)}$ & $H_0^{(1)}$ & $H_0^{(2)}$ & $H_0^{(1)}$ & $H_0^{(2)}$  \\ \cmidrule{3-8}
    \multirow{2}{*}{A} & Abdomen & \cellcolor{blue!20}0.001175 & 0.999863 & \cellcolor{blue!20}0.000002 & 1.000000 & \cellcolor{blue!20}0.000137 & 0.999992 \\
    & Chest & 0.314529 & 0.833847 & 0.500000 & 0.696381 & 0.226562 & 0.937500 \\
    \multirow{2}{*}{B} & Abdomen & \cellcolor{blue!20}0.046143 & 0.988770 & \cellcolor{blue!20}0.003693 & 0.999512 & 0.226562 & 0.937500 \\
    & Chest & 0.500000 & 0.812500 & \cellcolor{blue!20}0.001953 & 1.000000 & 0.500000 & 1.000000 \\
    \multirow{2}{*}{C} & Abdomen & 0.676197 & 0.500000 & \cellcolor{blue!20}0.031784 & 0.990395 & 0.500000 & 0.746094 \\
    & Chest & 0.171875 & 0.945312 & 0.312500 & 0.937500 & 1.000000 & 0.125000 \\
    \bottomrule
    \end{tabular}
\end{table*}

\end{document}